%% file: main.tex
\documentclass[journal]{IEEEtran}
\ifCLASSINFOpdf
\else
\fi
\usepackage{color}
\usepackage{listings}
\usepackage{multirow}
\usepackage{graphicx}
\graphicspath{ {images/} }
\usepackage[separate-uncertainty = true,multi-part-units = repeat]{siunitx}
\usepackage[linesnumbered,ruled,vlined]{algorithm2e}
\usepackage{amsmath}
\usepackage{caption}
\usepackage[inline]{enumitem}
\usepackage{changepage}
\usepackage[table,xcdraw]{xcolor}
\usepackage[T1]{fontenc}
\usepackage[utf8]{inputenc}
\usepackage[english]{babel}
\usepackage{vhistory}
\usepackage[hidelinks,bookmarks=true]{hyperref}
\usepackage{tabularx}
\usepackage{pdfpages}
\usepackage{pdflscape}
\usepackage{subcaption}

\begin{document}
%
\title{Evolving Learning Rate Optimizers for Deep Neural Networks}

\author{Pedro~Carvalho,
        Nuno~Lourenço,
        and Penousal~Machado
\IEEEcompsocitemizethanks{\IEEEcompsocthanksitem Pedro Carvalho, Nuno Lourenço and Penousal Machado are with the University of Coimbra, Department of Informatic Engineering, 3030-290 Coimbra, Portugal.\protect\\
E-mail: pfcarvalho@student.dei.uc.pt, \{naml, machado\}@dei.uc.pt}
\thanks{...}}


%



\maketitle

\begin{abstract}
Artificial Neural Networks (ANNs) became popular due to their successful application  difficult problems such image and speech recognition. However, when practitioners want to design an ANN they need to undergo laborious process of selecting a set of parameters and topology. Currently, there are several state-of-the art methods that allow for the automatic selection of some of these aspects. Learning Rate optimizers are a set of such techniques that search for good values of learning rates. Whilst these techniques are effective and have yielded good results over the years, they are general solution i.e. they do not consider the characteristics of a specific network. 

We propose a framework called AutoLR to automatically design Learning Rate Optimizers. Two versions of the system are detailed. The first one, Dynamic AutoLR, evolves static and dynamic learning rate optimizers based on the current epoch and the previous learning rate. The second version, Adaptive AutoLR, evolves adaptive optimizers that can fine tune the learning rate for each network eeight which makes them generally more effective. The results are competitive with the best state of the art methods, even outperforming them in some scenarios. Furthermore, the system evolved a classifier, ADES, that appears to be novel and innovative since, to the best of our knowledge, it has a structure that differs from state of the art methods.

\end{abstract}


%
\IEEEpeerreviewmaketitle

\section{Introduction}
\input{IEEEtran/1-Introduction}

\section{Background}

\input{IEEEtran/2-Background}
\section{AutoLR}
\input{IEEEtran/3-AutoLR}
\section{Experiments}
\input{IEEEtran/4-Experiments}

\section{Conclusion}
\input{IEEEtran/5-Conclusion}






%
\bibliographystyle{IEEEtran}
\bibliography{IEEEabrv,main.bib}
\begin{IEEEbiography}[{\includegraphics[trim={5cm 0 5cm 0},width=1in,height=1.25in,clip]{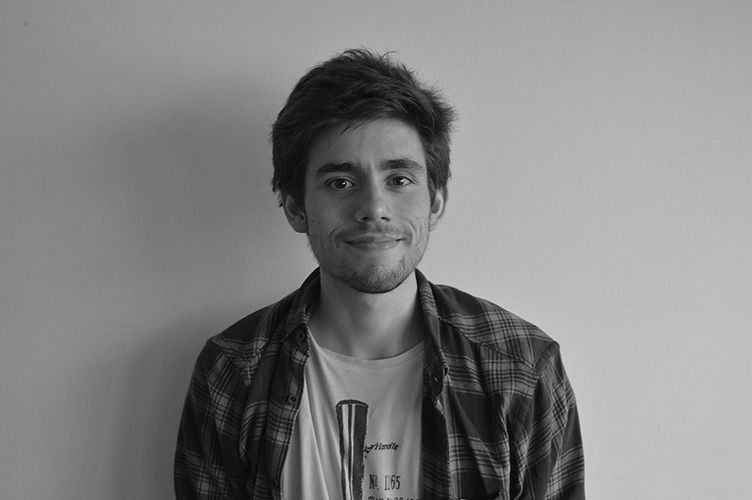}}]{Pedro Carvalho}
is a student and researcher at the University of Coimbra where he obtained a Master's Degree in Informatics Engineering. He is presently a grantee in Centre of Informatics and Systems of University of Coimbra (CISUC) working in the field of Evolutionary Machine Learning, specifically the evolution of learning rate optimizers. He is currently a PhD student in Information Science and Technology in the University of Coimbra.\end{IEEEbiography}
\begin{IEEEbiography}[{\includegraphics[width=1in,height=1.25in,clip,keepaspectratio]{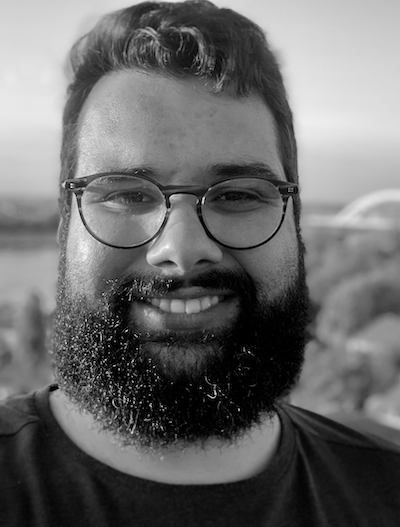}}]{Nuno Lourenço} is an Assistant Professor at the Department of Informatics Engineering of the University of Coimbra, where he obtained his PhD in Information Science and Technology in 2016. 
He is the current leader of the Evolutionary and Complex Systems (ECOS) group, and is a member of the Centre for Informatics and Systems of University of Coimbra (CISUC) since 2009. Formerly, he was appointed as a Senior Research Officer at the University of Essex in the United Kingdom. His main research interests are in the areas Bio-Inspired Algorithms, Optimisation and Machine Learning. 
He is the creator of Structured Grammatical Evolution, a genotypic representation for Grammatical Evolution, and participated in the proposal of a novel approach to automatically design Artificial Neural Networks (ANNs) using Evolutionary Computation called DENSER. 
He served as chair in the main conferences of the Evolutionary Computation field, namely EuroGP 2020 and EuroGP 2021 as program-chair, and PPSN 2018 and EuroGP 2019 as publication chair. He is member of the Programme Committee of GECCO, PPSN, EuroGP; member of the Steering Committee of EuroGP; and executive board member of SPECIES. He has authored or co-authored several articles in journals and top conferences from the Evolutionary Computation and Artificial Intelligence areas and he has been involved as a researcher in many national and international projects.
\end{IEEEbiography}
\begin{IEEEbiography}[{\includegraphics[width=1in,height=1.25in,clip,keepaspectratio]{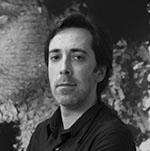}}]{Penousal Machado}
is an Associate Professor at the Department of Informatics of the University of Coimbra in Portugal. He is a deputy director of the Centre for Informatics and Systems of the University of Coimbra (CISUC), the coordinator of the Cognitive and Media Systems group and the scientific director of the Computational Design and Visualization Lab. of CISUC. His research interests include Evolutionary Computation, Computational Creativity, Artificial Intelligence and Information Visualization. He is the author of more than 200 refereed journal and
conference papers in these areas, and his peer-reviewed publications have been nominated and awarded multiple times as best paper. Until April 2020 his publications gathered over 2631 citations, an h-index of 24, and an i10-index of 61. He was the advisor of 7 PhD thesis and 39 MSc thesis, and is currently advising 7 PhD and 3 MSc thesis. He is also the chair of several scientific events, including, amongst the most recent, ICCC 2020, PPSN XV, and EvoStar 2016; member of the Programme Committee and Editorial Board of some of the main conferences and journals in these fields; member of the Steering Committee of EuroGP, EvoMUSART and Evostar; and executive board member of SPECIES. He is the recipient of several scientific awards, including the prestigious EvoStar Award for outstanding Contribution to Evolutionary Computation in Europe, and the award for Excellence and Merit in Artificial Intelligence granted by the Portuguese Association for Artificial Intelligence. Penousal Machado has been invited to perform keynote speeches in a wide set of domains, from evolutionary computation to visualization and art. His work was featured in the Leonardo journal, Wired magazine and presented in venues such as the National Museum of Contemporary Art (Portugal) and the “Talk to me” exhibition of the Museum of Modern Art, NY (MoMA).
\end{IEEEbiography}
\vfill
\end{document}

%% file: IEEEtran/1-Introduction.tex
Artificial Neural Networks (ANN) are an important part of modern Artificial Intelligence. These systems are adept at solving a variety of different tasks, showing a remarkable performance in the fields of computer vision \cite{Krizhevsky2017, Couprie2013}, medicine \cite{Esteva2019} and natural language processing \cite{Hochreiter1997, Lopez2017}.

ANN's design is loosely inspired by the workings of the human brain. Like their biological counterpart ANNs are comprised of several small units called neurons (or nodes). These units are interconnected and each connection has an associated value called a \textit{weight}. The weights determine the strength of the connections between neurons. When using an ANN for a specific problem there is a set of weight values that are most adequate to find a solution to he problem at hand. The process through which the correct set of weight for a given task are found is called \textit{training}. Training is paramount for the creation of an ANN that is able to correctly and consistently solve the target problem. The importance of this process led to extensive research into how ANNs should be trained and how to regulate this training. As a result there are currently several methodologies and/or hyperparameters used to tune the training process, one such hyperparameter being the \textit{learning rate} (LR).

Learning rate is a numeric value that scales changes made to the weights during training. The learning rate utilized has a profound impact on the effectiveness of training. A learning rate value that is too low may result in an excessively slow and ineffective training; a learning rate value that is too high may result in an erratic training process where the network never converges towards a set of stable weights.
There is also evidence that, as training progresses, the best course of action is to adjust the learning rate online as seen in \cite{Senior2013}.
The high impact of the learning rate in training efficiency motivated the research community to find a variety of ways to optimize the use of this parameter. These learning rate optimizations solutions will be referred to as \textit{LR optimizers} throughout the rest of this work. There are several LR optimizers with varying levels of complexity and effectiveness \cite{rmsprop, Kingma2014, NESTEROV1983, bottou-98x}. One aspect most LR optimizers share, however, is their generality. Since training is ubiquitous across most applications of ANNs, LR optimizers are designed to be effective regardless of the problem or network in question.

This general approach has led to the creation of LR optimizers that are effective and easy to use, but it also raises the question: Can LR optimizations be pushed further if we specialize these mechanisms to the problem in question? 

To answer this question we must first establish a way to specialize LR optimizers for a specific problem. Since ANNs are comprised of many interdependent components and parameters it is impossible for a human to understand all the dimensions required for manual specialization. It is possible, however, to use a search algorithm to perform this specialization automatically. Evolutionary algorithms (EA) are the most suited for this task; these heuristic algorithms are able to navigate complicated problem spaces efficiently through biologically inspired procedures (e.g. crossover, mutation, selection).
Using an EA it is possible to test several different optimizers and combine the best performing ones to achieve progressively better results. These evolved optimizers can then be compared with standard, man-made optimizers to assess the benefits of specialization.

In this article we present and evaluate a framework that is able to evolve optimizers for specific machine learning problems. Furthermore, the resulting evolved optimizers are benchmarked against state of the art standard optimizers. Finally, the applicability of evolved optimizers in general cases is also tested empirically. The results suggest that the evolved optimizer are able to compete with state of the art standard solutions, outperforming them in some test scenarios. Furthermore, some evolved optimizers exhibit novel behavior not found in the literature.

The structure of this paper is the following:
\begin{itemize}
    \item Section 2 gives a historical background on the standard LR optimization techniques discovered over the years.
    \item Section 3 describes AutoLR, the framework created to evolve LR optimizers on a general level. Sections 4 and 5 are in-depth explanations of the important components of the framework.
    \item Section 6 outlines the experiments performed to assess the viability of AutoLR. This section also presents and analyses the results obtained in our experimentation.
    \item Section 7 reviews the work presented in this article and summarize our contributions.
\end{itemize}

%% file: IEEEtran/2-Background.tex
ANNs are configured by a set of hyperparameters. One such parameter is the learning rate, this parameter scales the changes made to the network's weights during training. This parameter has a profound effect on the effectiveness of the training and the network's subsequent performance. 

After each training epoch the system compares the output of the network with the expected output and calculates the error. Based on this error, back propagation \cite{Rumelhart1986} is used to calculate the changes that should be made to each individual weight (known as the gradient). There are several optimizers that take the gradient and use it the change the weights. The term learning rate refers to a value that is used in many of these optimizers to scale the gradient before it is applied to the weights. 

The original learning rate optimizer, SGD \cite{bottou-98x}, simply sets the new weight ($w_{t}$) to be the difference between the old weight ($w_{t-1}$) and the product of the learning rate ($lr$) with the gradient ($\nabla l$), shown in Equation~\ref{eq:sgd}.

\begin{equation}
w_{t} = w_{t-1} - \text{lr} * \nabla l(w_{t-1})
\label{eq:sgd}
\end{equation}

Traditionally, a single learning rate value is used for the entirety of training. In this case, all the tuning must be done before the training starts. The issue with this approach is that one is often forced to rely of experience and trial-and-error to find a good static learning rate. This task becomes even harder when other hyper parameters are considered. Since the hyperparameters in ANNs are inter-dependent there is no guarantee the learning rate remains adequate once other parameters are adjusted. 

These limitations led to the creation of dynamic learning rates. Dynamic learning rates varies as training progresses. A most common approach is to start with a high learning rate that decreases as training progresses. This approach is effective because it covers a range of different learning rate values and decreases the value over time which is theoretically desirable. Dynamic approaches are commonly used \cite{suganuma2017, he2016} as they typically outperform static learning rates \cite{Senior2013}. 

Dynamic learning rates are still limited because they have no knowledge of what is happening throughout the training process. The optimizers can change the learning rate based on the training epoch but not based on changes in the gradient. This lead to the development of the most sophisticated LR optimizers; adaptive LR optimizers.

Adaptive learning rate optimizers are variations of SGD that use additional functions to adjust the learning rate for each weight individually. Adaptive optimizers are able tune the different learning rate for individual weights through the use of auxiliary variables that are tracked for each weight. The simplest adaptive optimizer is the momentum optimizer \cite{Sutskever2013}. Equation \ref{eq:momentum} shows this variation of SGD. In this solution the auxiliary variable is a momentum term ($x_t$) that increase the size of adjustments made to weights that keep changing in the same direction. This term is accompanied by two constants. Learning rate (lr) is responsible for directly scaling the gradient. The momentum constant (mom) takes a value between 0 and 1 that dictates how strong the effect of the momentum is.
\begin{equation}
\begin{aligned}
x_{t} &= \mathrm{mom} * x_{t-1} - \mathrm{lr} * \nabla l(w_{t-1})\\
w_{t} &= w_{t-1} + x_{t}        
\end{aligned}
\label{eq:momentum}
\end{equation}

A variation of the momentum optimizer, known as Nesterov's momentum \cite{NESTEROV1983} is presented in Equation~\ref{eq:nesterov}. Nesterov's momentum varies from the original because the gradient is calculated for the weight plus the momentum term. The optimizer is able to look-ahead and make corrections to the direction suggested by the momentum. This is beneficial because the momentum term is slow to change and can hinder the training process as a result.

\begin{equation}
\begin{aligned}
x_{t} &= \mathrm{mom} * x_{t-1} - \mathrm{lr} * \nabla l(w_{t-1} + mom * x_{t-1})\\
w_{t} &= w_{t-1} + x_{t}        
\end{aligned}
\label{eq:nesterov}
\end{equation}

RMSprop is an unpublished optimizer that divides the learning rate by a moving discounted average of the changes made to the weights. Practically speaking this optimizer will decrease the learning rate when a weight is changing rapidly and increase it when the weight stagnates. This learning rate annealing simultaneously helps the weights converge and prevents them from stagnating. In Equation~\ref{eq:rmsprop} $x_t$ is the moving average term and $\rho$ is the exponential decay rate used for this same average. The root of the moving average is then used in $w_t$ to scale the learning rate and gradient.
        \begin{equation}
        \begin{aligned}
      x_{t} &= \rho x_{t-1} + (1 - \rho) \nabla l(w_{t-1})^2\\
      w_{t} &= w_{t-1} - \frac{lr * \nabla l(w_{t-1})}{\sqrt{x_{t}} + \epsilon}        
        \end{aligned}
    \label{eq:rmsprop}
  \end{equation}

The final optimizer we will be discussing is Adam \cite{Kingma2014}. In the original work Adam is shown to outperform the other optimizers present on several problems. Adam is similar to RMSprop but it attempts to correct the bias that comes with starting the moving average at 0. The new term $z_t$ is used to correct this bias. Adam also uses $\frac{x_{t-1}}{\sqrt{y_{t-1}}}$ to calculate a range where it expects the gradient to remain consistent. In Equation~\ref{eq:adam} $x_t$ and $y_t$ are both moving averages; consequently, $\beta_1$ and $\beta_2$ are exponential decay rates for the averages (similar to $\rho$ in Equation~\ref{eq:rmsprop}).

\begin{equation}
\begin{aligned}
x_{t} &= \beta_1 x_{t-1} + (1 - \beta_1) \nabla l(w_{t-1})\\
y_{t} &= \beta_2 y_{t-1} + (1 - \beta_2) \nabla l(w_{t-1})^2\\
z_{t} &= lr * \frac{\sqrt{1 - \beta_2^t}}{(1 - \beta_1^t)}\\
w_{t} &= w_{t-1} -  z_{t} * \frac{ x_{t}}{\sqrt{y_{t}} + \epsilon}
\end{aligned}
\label{eq:adam}
\end{equation}

%% file: IEEEtran/3-AutoLR.tex
AutoLR \cite{Carvalho2020} is an open source \cite{autolr_repo} framework developed to evolve LR optimizers. In the context of this work this framework is utilized to assess the benefits of creating LR optimizers specialized for specific machine learning tasks.

AutoLR should be used in two phases. In the first phase, evolution is performed using Dynamic Structured Grammatical Evolution \cite{lourencco2018structured}. The evolution stage requires the following components: 

\begin{itemize}
    \item Engine
    \item Grammar
    \item Fitness Function
    \item Evolutionary Parameters
    \item Machine Learning Task
\end{itemize}

Each of these components is described in depth in the next section.

Once the evolution phase is complete it is recommended that an additional step is taken to validate the quality of the LR optimizers produced. We call this phase benchmark. In the benchmark phase the evolved optimizers are compared with their standard counterparts on the task used during evolution.

\section{Evolution}
\subsection{Engine}
In the context of this work, LR optimizers will be executable computer code. Certain approaches are best suited to evolve this type of individuals, namely Genetic Programming (GP) and Grammatical Evolution \cite{ONeill2001} (GE). We opted to use GE because the optimizers have common structure among them. GE enables the enforcement of this structure through an understandable and easily editable grammar. We chose to used a variation of GE in AutoLR, Dynamic Structured Grammatical Evolution (DSGE) \cite{lourencco2018structured}. 

In traditional GE, the genotype is a single list of integers that are translated into solutions using the grammar. Using a single list genotype handicaps the approaches locality since the meaning of a segment of the genotype depends on what precedes it. In other words, small changes in genetic material can lead to very different individuals in traditional GE and this makes evolution arduous. In DSGE, the genotype  holds one list of integers for each production in the grammar. This change improves locality since the genetic material is closely tied to a specific part of the grammar. With improved locality DSGE is able to enact a more efficient evolution which motivated its use in AutoLR.

The original DSGE is implemented in Python. This language also has extensive support for machine learning (through the Tensorflow \cite{tensorflow} and Keras \cite{keras} libraries) and the whole framework was built in it as a result.

\subsection{Grammar}
Adequate grammar design is paramount for successful evolution \cite{Nicolau2018, nicolau2012}. The grammar used determines the type of LR optimizers the system is able to create. While grammar design is largely subjective, there are a few guidelines that should be taken into account during this process. 

The first step when creating a grammar for AutoLR is ensuring the system is able to reproduce some of the standard LR optimizers. Our objective when using this tool is to go beyond the standard approaches; nevertheless the presence of these optimizers is important as it ensures the evolutionary process is able to produce functional individuals that can guide evolution. Additionally, the presence of standard optimizers in the evolutionary phase serves as a sanity check that guarantees the system is creating functions that will help solve the target problem.

Since evolutionary algorithms demand a large quantity of computational resources, one must also take measures to prevent the creation of a problem space that is too complex. The grammar utilized effectively defines the search space the algorithm will be navigating. Consequently it is important that, when designing the grammar, we consider the complexity of the resulting search space. Ideally, the search space such be concise enough that the algorithm is able to explore it effectively with the resources available but complex enough that the resulting individuals are adequate.

Finally, it is important to consider that grammars can be biased in order to favor the creation of certain individuals. Biasing the grammar can be an efficient way of accelerating the progress of the search process. In an unbiased grammar there is a risk that the number of useful individuals is so small in comparison to the number of all possible individuals that the evolutionary algorithm is simply unable to progress. In AutoLR we are using the evolutionary algorithm to create solutions with a specific task. Finding the common ground between these desirable solutions and biasing the grammar towards this common ground decreases the number of invalid individuals without compromising the system's ability to innovate.

\subsection{Fitness Function}
Developing a fitness function in this context can be a challenging task. The most obvious and necessary component of the fitness function is that the generated individual most be used to solve the machine learning task. It is recommended however that additional measures are taken to ensure the fitness value is an accurate measure of the solutions actual performance.

In order to ensure that the evolved optimizer is helping with the resolution of the machine learning task chosen it is important to consider the possibility of overfitting. Using the score obtained in training directly as the fitness is not ideal since the evolved individual might be unable to generalize beyond the training data. To address this one should take the network trained using the evolved optimizer and test it on a new dataset; the score obtained in this second dataset is a more accurate representation of the solution's true fitness. It must also be noted that due to the nature of AutoLR it is possible that the evolved optimizers will implicitly become optimized towards the second dataset as well. Consequently, the user should always keep a third dataset, that is never used during the evolutionary process, for benchmarking purposes.

There are several stochastic components to the machine learning process; this means that when testing the same solution twice it is expected that the results will be slightly different. In some situations we might find solutions that are so inconsistent that the result of a single trial is not an accurate measure of their true fitness. In these cases it is paramount that the fitness function is able to work around these inconsistencies.
In order to address these it is necessary to increase the number of trials performed on a single solution. However since both machine learning and evolutionary algorithms are resource intensive approaches, it is desirable that this increase in trials is kept to a minimum. 

The number of trials can be minimized through additional mechanisms. It is important for LR optimizers to be able to produce good results consistently, based on this fact we found two mechanisms that can be used to reduce the number of trials. The first mechanism exploits the fact that a large number of individuals produced by the evolutionary algorithm are not functional optimizers; multiple trials are only necessary when we are looking to discern between the better of two optimizers. Defining a \textit{minimum acceptable score} creates a threshold that a candidate solution must surpass in order to warrant repeat trials. Once a solution has been evaluated several times it is also necessary to devise a way to consolidate all the results into a single fitness value. While using the average is the most straightforward answer we found that using the \textit{minimum score across all trials} further incentivizes the system to produce solutions that are consistently good.

\subsection{Machine Learning Task}

Since the machine learning part of the system is isolated, AutoLR can be applied to virtually any machine learning task. The only requirement for integration is that the task chosen is capable of utilizing the generated optimizers. To be the best of our knowledge, \textit{any machine learning task where training is performed can benefit from AutoLR}.

In order to solve a machine learning problem we typically require a neural network and a dataset. The network and dataset chosen will be used by the framework through the fitness function. Both the network and data used affect the cost of experiments; bigger networks and data will slow down the training process and, as a result, the evolutionary process.
In this work all the experiments were performed for a single combination of network and dataset at a time. Not all optimizers produced in this way will be able to generalize to other task. It is possible to evolve optimizers for several networks or datasets (or even both) but the efficacy of this approach is unproven.

While the AutoLR's versatility enables it to evolve optimizers for most problems, it is important to consider the trade-offs that come with this tool. While AutoLR has several parameters that control the cost of its trials, we found that in order to produce a solution that is significantly better than standard approaches requires a considerable amount of resources. It is important to consider that while surpassing standard optimizers is a costly endeavor, producing evolved optimizers of comparable performance is much less so. In the field of LR optimizers, ideas from inferior optimizers are frequently reused and expanded upon by more sophisticated approaches. Consequently, optimizers created by AutoLR can theoretically showcase a way to improve the standard approach in use.

\section{Benchmark}

In the previous section we presented several guidelines that help the reader perform successful evolutionary runs. It is important to highlight that many of these techniques aim to reduce the time spent assessing which individual's fitness. It is expected that, in the process of speeding up the fitness function, the fitness values produced lose some of their accuracy. In other words, the fitness function used during the evolution phase yields an approximation of the solution's real ability to solve the task. As a result, it is important that additional steps are taken in order evaluate the evolved optimizers' true fitness and how they benchmark against standard approaches. This step of the experiment is called the \textit{benchmark phase}.

In the benchmark phase we are only going to be evaluating solutions a negligible amount of times compared to the evolution. Consequently, it is recommended to make the fitness function as accurate as possible, even if it becomes significantly slower as a result. In this phase, it is also possible to use all available data to make the most accurate assessment possible. 

When comparing evolved optimizers with the standard ones it is also important to avoid testing on any of the data present during the evolutionary process. Since the fitness function effectively guides the evolutionary process it is possible that the solutions produced have become attached to the data used. If benchmarking is performed on this same data the evolved optimizers will have an inherent advantage, compromising the results.

%% file: IEEEtran/4-Experiments.tex
In order to analyze and validate the performance of our proposal several experiments were performed using the AutoLR framework. The object of these experiments was to evolve and benchmark optimizers of different classes.

\subsection{Dynamic AutoLR}
Dynamic AutoLR (DLR) \cite{Carvalho2020} is an implementation of AutoLR designed to evolve static and dynamic optimizers. The union of these two types of optimizers is also known as \textit{learning rate schedulers} or \textit{learning rate policies}.

This implementation of the system was created to validate the hypothesis that optimizers evolved through the framework are useful and comparable with standard approaches. Learning rate schedulers benefit from some characteristics that make them the ideal for a proof of concept. All LR optimizers are mathematical functions, but compared to the more complex adaptive optimizers, learning rate schedulers only take two inputs (previous learning rate and epoch) and produce a new learning rate from those. Since policies have access to such limited information, the behaviors they produce can be approximated with decisions trees. The number of operations used in decision trees is much smaller than the one used for mathematical functions. As a result, the grammars used to produce decision trees can be plainer and easier to navigate for the evolutionary algorithm. 

The grammar and design decisions made for DLR are explained in detail in other works \cite{Carvalho2020}. The most important characteristic of the grammar used is that the resulting individuals are comprised of a series of if-the-else constructs. These individuals evaluate a series of conditions using the current epoch and learning rate and return a new learning rate for the next epoch. It is important to note that these types of individuals do not replicate standard dynamic optimizers directly. Instead, these decision trees are able to approximate methods such as decaying \cite{Mishra2019} and cyclical \cite{smith2017cyclical} learning rates. The parameters used for DLR can be found in \cite{Carvalho2020}.

These experiments were also used to assess the impact that early stop has on the evolutionary process. Experimentation in DLR is separated into two scenarios: scenario I trains the network for 100 epoch and uses an early stop mechanism; scenario II trains the network for 20 epochs and no early stop.

Image classification was the problem chosen as there is a vast backlog of research on the topic. The presence of this backlog enabled us to use proven models and datasets allowing our focus to be on the tuning of the evolutionary component.

\begin{figure*}
    \includegraphics[width=\textwidth,keepaspectratio]{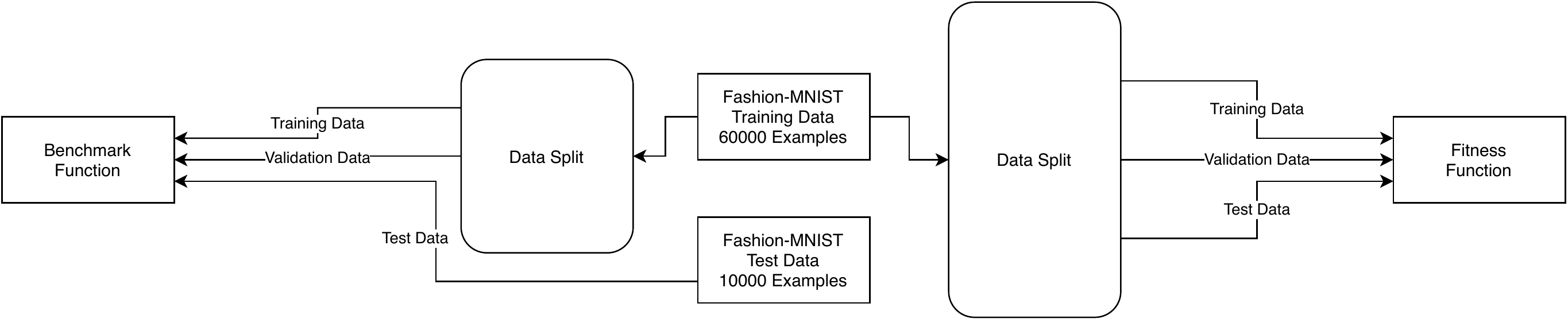}
    \caption{Difference in data used for evolution (fitness function) and benchmark (benchmark function).}
    \label{fig:data_split} 
\end{figure*}

When choosing a dataset it is important to consider two aspects. The difficulty of the dataset must be high since it is crucial for the system to be able to differentiate between mediocre and great optimizers. If the chosen problem is too easy, all competent optimizers will obtain indistinguishably good scores. The other aspect that must be considered is the size of each example. Bigger images take longer to classify; thousands of images are classified in each training epoch and, since training itself is performed thousands of times throughout the experiment, the resulting slow down is massive. 
We found that Fashion-MNIST is a good balance between a dataset that is too easy (such as the regular MNIST) and one that is too complex (such as CIFAR-10). In evolution, 7000 instances were used for training, 1500 for validation and another 1500 for testing. Fashion-MNIST itself is separated into a training and a testing dataset. All the instances present in evolution are part of the training set. As a result, the evolutionary process never comes into contact with Fashion-MNIST's test examples. This is extremely important since this test data can be used to make an unbiased and fair benchmark later. Figure~\ref{fig:data_split} illustrates how the data is split between evolution and benchmark. During evolution, data augmentation was also used.

The neural network architecture used in DLR comes from DENSER \cite{assuncao2018gpem}. This architecture is unique because it was created by an evolutionary algorithm, similar to the optimizers AutoLR produces. This characteristic makes this network an interesting one for this experiment. Since this architecture is quite complex it is more likely that there is room for the LR optimizer to be specialized. Additionally, DENSER evolves architectures with a \textbf{static learning rate of 0.01}; this policy is a good baseline for the benchmark phase since we know it works well for the network.

The final aspect of evolution that must be discussed is the fitness function. The evolutionary runs for DLR are short; as a result there is no need to utilize any of the possible optimizations previously discussed. The function used is shown in Algorithm~\ref{dlr_fitness_function}. It is important to highlight that the learning rate policy never comes into contact with the test data directly, this prevents the evolutionary algorithm from overfitting policies to the data.

\begin{algorithm}
\SetKwInOut{Params}{params}
\Params{network, learning\_rate\_policy, training\_data, validation\_data, test\_data}
trained\_network $\leftarrow{}$ train(network, learning\_rate\_policy, training\_data, validation\_data)\;
fitness\_score $\leftarrow{}$ get\_test
\_accuracy(trained\_network, test\_data)\;
\Return fitness\_score\;

\caption{Simplified version of the fitness function used in DLR}
\label{dlr_fitness_function}
\end{algorithm}

For each of the sets of 10 evolutionary runs, the best evolved schedulers was selected for benchmarking. Before moving onto the results it is important to analyze the shape of these evolved policies. By comparing the shape of the evolved solutions to the standard ones we can learn about the similarities between the two approaches.


Policy A was produced in scenario I, where the early stop mechanism is active. Policy A alternates between three learning rate values: a minimum, an intermediate one and a maximum; in that order. This policy appears to use some of the ideas present in the cyclical learning rates found in \cite{smith2017cyclical}. The policy changes between these value every epoch. The policies presented in that work vary the learning rate in a more controlled way, nevertheless we find the fact the system was able to produce similar behavior with no prior knowledge of the method is noteworthy.

Policy B is a static learning rate policy for the first 60 and produces erratic behavior afterwards. It is important to keep in mind, however, that since this policy was evolved in scenario II the training performed only lasted for 20 epochs. As a result, all learning rate changes after epoch 20 did not manifest during the evolutionary process.

Three benchmarks were performed for DLR:
\begin{itemize}
    \item Benchmark I - Replicates the conditions of Scenario I. Network is trained for  \textbf{100 epochs with the early stop mechanism}. 
    \item Benchmark II - Replicates the conditions of Scenario II. Network is trained for only \textbf{20 epochs, with no early stop}.
    \item Benchmark III - Designed to assess the peak performance of the optimizers. Network is trained for \textbf{100 epochs, with no early stop}.
\end{itemize}

In order to benchmark the evolved policies it is necessary to establish a baseline. The network architecture we are using was evolved using a static learning rate of 0.01; this policy will serve as the baseline since we know it works well with the network. Since this policy is the baseline it will be tested in all three benchmarks. The evolved optimizers will only be tested in their native conditions (benchmark I for policy A, benchmark II for policy B).

\subsubsection{Results}
Each policy was tested on each of their scenarios five times. The results of these trials are shown in Table~\ref{tab:dlr_test_results}.


\begin{table}[ht]
\centering
\resizebox{\columnwidth}{!}{%
\begin{tabular}{|
>{\columncolor[HTML]{EFEFEF}}c |
>{\columncolor[HTML]{EFEFEF}}c |c|c|c|}
\hline
\multicolumn{2}{|c|}{\cellcolor[HTML]{EFEFEF}}                             & \multicolumn{3}{c|}{\cellcolor[HTML]{EFEFEF}\textbf{Policy}}                \\ \cline{3-5} 
\multicolumn{2}{|c|}{\multirow{-2}{*}{\cellcolor[HTML]{EFEFEF}\textbf{Benchmark}}} &
  \cellcolor[HTML]{EFEFEF}\textbf{A} &
  \cellcolor[HTML]{EFEFEF}\textbf{B} &
  \cellcolor[HTML]{EFEFEF}\textbf{Baseline} \\ \hline
\cellcolor[HTML]{EFEFEF}                             & \textbf{Validation} & 75.09 $\pm$ 16.79\%           & \cellcolor[HTML]{343434} & \textbf{85.87 $\pm$ 0.25\%} \\ \cline{2-3} \cline{5-5} 
\multirow{-2}{*}{\cellcolor[HTML]{EFEFEF}\textbf{I}} &
  \textbf{Test} &
  69.24 $\pm$ 24.07\% &
  \multirow{-2}{*}{\cellcolor[HTML]{343434}} &
  \textbf{85.01 $\pm$ 0.36\%} \\ \noalign{
\hrule height 1pt
}%
\cellcolor[HTML]{EFEFEF}                             & \textbf{Validation} & \cellcolor[HTML]{343434} & 85.42 $\pm$ 0.87\%             & \textbf{85.55 $\pm$ 0.38\%} \\ \cline{2-5} 
\multirow{-2}{*}{\cellcolor[HTML]{EFEFEF}\textbf{II}} & \textbf{Test}       & \cellcolor[HTML]{343434} & \textbf{84.82 $\pm$ 0.71\%}    & 84.39 $\pm$ 0.22\%          \\ \noalign{
\hrule height 1pt
}%
\cellcolor[HTML]{EFEFEF}                             & \textbf{Validation} & \textbf{89.38 $\pm$ 0.39\%}    & 89.13 $\pm$ 0.33\%             & 88.85 $\pm$ 0.22\%          \\ \cline{2-5} 
\multirow{-2}{*}{\cellcolor[HTML]{EFEFEF}\textbf{III}} & \textbf{Test}       & \textbf{88.68 $\pm$ 0.22\%}    & 86.58 $\pm$ 1.47\%             & 87.47 $\pm$ 0.37\%          \\ \hline
\end{tabular}
}
    \caption{Accuracy of the evolved policies (A \& B) on their evolutionary environment (benchmark 1 \& 2 respectively) and scenario 3 (representative of max policy performance), compared with the baseline policy.}
    \label{tab:dlr_test_results}
\end{table}

Benchmark I compares policy A with the baseline policy using a training of 100 epochs with early stop. It is interesting to note that results show that, despite being evolved in these same conditions, policy A performed worse than the baseline. The variation in results obtained by the evolved policy in this benchmark were extremely large in comparison to all other results so we performed an analysis of each trial individually. The reason policy A under-performs in this benchmark is because the balance between the policy and the early stop mechanism is delicate; in some runs an early stop is triggered within the first epochs of training, leading to a poor performance. When the policy is able to overcome the first epochs without activating the mechanism, it is able to outperform the baseline.

Benchmark II compares policy B with the baseline policy using 20 epochs of training and no early stop. The results in this benchmark are straightforward. Policy B obtains a better test score on average than the baseline. It should be noted that the evolved policy outperforms because it does not overfit as much as the baseline; this is evidenced by the difference in accuracy when moving from validation to test data. Finally it must be noted that policy B is more inconsistent than the baseline, despite showing better results on average.

Benchmark III compares both of the evolved policies with the baseline, training for 100 epochs with no early stop. This benchmark is designed to assess if the upper limit of the evolved policies is superior to that of standard solutions. Policy A produces the best and most consistent test scores in this benchmark. It is interesting to note that, despite being evolved using the early stop mechanism, this policy benefits greatly from its removal. Policy B is able to achieve results similar to policy A in validation but suffers a loss in performance when moved to testing due to overfitting. This loss is expectable since the policy evolved a static learning rate to learn in 20 epochs; when moved into 100 epochs, this same learning rate is unable to make the most of the extra resources. Finally, the baseline seems to learn slowly and accurately in this benchmark since it is weaker in validation than the evolved policies but is hardly affected by overfitting.

We found that these results show that AutoLR has potential to create interesting and competitive optimizers. DLR only works with a small subset of all LR optimizers that are not commonly regarded as the best approaches available. These facts motivated the development of a second version of the system that deals with the more complex adaptive optimizers.

\subsection{Adaptive AutoLR}
Adaptive AutoLR (ALR) is an implementation of the AutoLR framework to evolve adaptive optimizers. The key difference between this version of the system and DLR is that the grammar (and resulting solutions) are far more complex. When dealing with learning rate policies the function always returns a learning rate; this means that the worst case scenario is that the learning rate chosen is inadequate. Adaptive optimizers are a small group of functions that, when combined, are supposed to adjust the weight of the network based on the gradient. This definition is much broader than that of a learning rate policy.
This added complexity demands the number of evaluations used in evolution be increased by a few orders of magnitude. The cost of experiments is far greater in ALR and several changes made from DLR are motivated by this fact. 

Another consequence of the broad definition of adaptive optimizer is that the majority of possible solutions are not able to train the network. The easiest way to counteract this issue would be through a restrictive grammar that limits the types of optimizers that can be evolved.
In this work we are interested in promoting optimizers that deviate from the standard approaches as much as possible and, as a result, will be avoiding such restrictions. This in turn means that the presence of individuals that can guide the evolutionary process must be guaranteed and incentivized. 
The full grammar used for ALR cannot be included due to space restrictions but an abridged version is presented in Figure~\ref{fig:alr_grammar} (full version can be found in \cite{autolr_repo}). 
When designing the grammar for ALR reproducibility was a high priority i.e. the grammar is capable of producing a large amount of standard adaptive optimizers (e.g. SGD, RMSprop, Adam). The resulting individuals are comprised of 4 functions, named: $x\_func$, $y\_func$, $z\_func$ and $weight\_func$. Functions $x$ through $z$ work as the auxiliary functions found in standard adaptive optimizers; these functions have an associated result that is carried between epochs (e.g. $x_t$). $weight\_func$ is the weight update function, the result of this function will the used as the weight for the next epoch.
Figure~\ref{fig:alr_grammar} only shows the productions related to the $x$ and $weight$ functions; $y$ and $z$ have the same productions as $x$ except for a few exceptions. The auxiliary functions are executed in the following order: $x$ then $y$ and finally $z$. This order is important because each auxiliary function has access to the result of those that precede it in the order, in their $terminal$ production. Naturally, the $weight$ function is executed after all the auxiliaries and has access to all their results. There are several productions in this grammar that are identical. These productions cannot be combined because we want SGE to keep the genotype of each individual function segregated. The operations chosen for the $func$ productions were all chosen for their presence in standard adaptive optimizers. The constants used in the $const$ productions follow a sigmoidal distribution between 0 and 1. This distribution is used instead of a linear one because adaptive solutions frequently utilize values very close to 1 as decaying factors. 
The gradient cannot be directly used by the weight function; this encourages the use of auxiliary functions. Auxiliary functions' terminals are also biased so the gradient is picked more often, speeding up the discovery of functioning optimizers. The auxiliary functions naturally accumulate their value which makes it harder for them to clearly relay the gradient to the weight function. While this cumulative property is desirable, it makes the initial discovery of functioning optimizers very difficult. The $expr$ productions are included to allow the system to easily discover out to remove the cumulative property from the function.

Since reproducibility was a top priority when designing the grammar it is exceedingly important to highlight the behaviors that cannot be evolved. In order to keep performance as quick as possible the auxiliary functions are calculated using Tensorflow's gradient descent training operation. This decision makes it impossible to calculate modified gradients such as the one used in Nesterov's momentum since, to be the best of our knowledge, it is not possible to obtain this information without using the corresponding training operation. Additionally, some adaptive optimizers allow the use of different initial values for their auxiliary variables. In ALR, all auxiliary variables are initialized as 0; enabling the evolution of initial values could hurt the development of the functions so we opted not to do so.

\begin{figure}
    \centering
    \begin{small}
    \begin{align*}
        {<}\text{start}{>}::= & \, \text{ x\_func, y\_func, z\_func, weight\_func =} \\
        & {<}\text{x\_expr}{>}\text{, }  {<}\text{y\_expr}{>}\text{, }
        {<}\text{z\_expr}{>}\text{, } {<}\text{weight\_expr}{>} \\ 
        {<}\text{x\_expr}{>}::= & \, \text{ add(x, }{<}\text{x\_update}{>}\text{) } \, | \, \text{ }{<}\text{x\_update}{>} \\ 
        {<}\text{x\_update}{>}::= & \, \text{ }{<}\text{x\_func}{>}\text{ } \, | \, \text{ }{<}\text{x\_terminal}{>} \\ 
        {<}\text{x\_func}{>}::= & \, \text{ negative(}{<}\text{x\_expr}{>}\text{) } \, \\ 
        & | \, \text{ subtract(}{<}\text{x\_expr}{>}\text{, }{<}\text{x\_expr}{>}\text{) } \, \\
        & | \, \text{ multiply(}{<}\text{x\_expr}{>}\text{, }{<}\text{x\_expr}{>}\text{)  } \, \\
        & | \, \text{ pow(}{<}\text{x\_expr}{>}\text{, }{<}\text{x\_expr}{>}\text{) } \, \\
        & | \, \text{ square(}{<}\text{x\_expr}{>}\text{) } \, \\
        & | \, \text{ divide\_no\_nan(}{<}\text{x\_expr}{>}\text{, }{<}\text{x\_expr}{>}\text{) } \, \\
        & | \, \text{ add(}{<}\text{x\_expr}{>}\text{, }{<}\text{x\_expr}{>}\text{) } \, | \, \text{ sqrt(}{<}\text{x\_expr}{>}\text{) } \, \\
        {<}\text{x\_terminal}{>}::= & \, {<}\text{x\_const}{>} \, | \, \text{ x } \, | \, \text{ grad } \, | \, \text{ grad} \\ 
        {<}\text{x\_const}{>}::= & \, \text{ 4.53978687e-05 } \, | \, \ldots \, | \, \text{ 9.99954602e-01} \\
        \ldots \\ 
        {<}\text{weight\_expr}{>}::= & \, \text{ }{<}\text{weight\_func}{>}\text{ } \, | \, \text{ }{<}\text{weight\_terminal}{>} \\ 
        {<}\text{weight\_func}{>}::= & \, \text{ negative(}{<}\text{weight\_expr}{>}\text{) } \, | \, ... \, \\
        {<}\text{weight\_terminal}{>}::= & \, {<}\text{weight\_const}{>} \, | \, \text{ x } \, | \, \text{ y } \, | \, \text{ z} \\ 
        {<}\text{weight\_const}{>}::= & \, \text{ 4.53978687e-05 } \, | \, \ldots \, | \, \text{ 9.99954602e-01} \\
\end{align*}
\end{small}
    \caption{CFG for the optimisation of learning rate optimizers.}
    \label{fig:alr_grammar}
\end{figure}

The fitness function suffered some changes in order to improve the consistency of the evolved solutions and evaluation speed. The updated function is shown in Algorithm~\ref{fitness_function_alr}. The optimizers are now trained and tested up to 5 five times. After each of these trials the system checks if the test score is above a minimum acceptable score. If the optimizer does not hit the threshold then the rest of the trials are canceled. This decreases the amount of resources spent on evaluating mediocre optimizers.
Furthermore, the worst score out of the 5 trials is used as the optimizers fitness; selecting the worst score penalizes inconsistent solutions.

The network architecture was also changed to favor a faster training. The network used in ALR can be found in \cite{keras_mnist}. This network is compatible with the Fashion-MNIST dataset and much faster to train due to a reduced number of weights. We also found that the data augmentation was slowing down the training process significantly so it was replace with additional data.

\begin{algorithm}
\SetKwInOut{Params}{params}
\Params{network, learning\_rate\_optimizer, training\_data\_groups, validation\_data, test\_data, trial\_number, }
minimum\_acceptable\_score $\leftarrow{}$ 0.8\;
fitness\_score $\leftarrow{}$ 1.0\;
trial\_count $\leftarrow{}$ 0\;
\While{trial\_count $<$ trial\_number}{
    trained\_network $\leftarrow{}$ train(network, learning\_rate\_optimizer, training\_data\_groups$[$trial\_count$]$, validation\_data)\;
    trial\_test\_score $\leftarrow{}$ get\_test\_accuracy(trained\_network, test\_data)\;
    \If{
        trial\_test\_score $<$ fitness\_score}{
        fitness\_score $\leftarrow{}$ trial\_test\_score\;
        }
    \If{
        trial\_test\_score $<$ minimum\_acceptable\_score}{
        \Return fitness\_score\;
        }
    trial\_count $++$\;
}
\Return fitness\_score\;

\caption{Simplified version of the fitness function used to evaluate optimizers in ALR}
\label{fitness_function_alr}
\end{algorithm}

\begin{table}[ht!]
    \centering
    \begin{tabular}{c | c}
        \textbf{SGE Parameter} & \textbf{Value} \\ \hline
        Number of runs & 9 \\ 
        Number of generations & 1500 \\  
        Number of individuals & 20 \\
        Tournament size & 5 \\
        Mutation rate &  0.15\\ \hline
        \textbf{Dataset Parameter} & \textbf{Value} \\ \hline
        Training set &  53000 instances from the training\\
        & 10600 instances per trial \\
        Validation set & 3500 instances from the training \\ 
        Test set &  3500 instances from the training \\ \hline
        \textbf{Early Stop} & \textbf{Value} \\ \hline
        Patience & 5 \\
        Metric & Validation Loss \\
        Condition & Stop if Validation Loss does not\\
        & improve in 5 consecutive epochs \\ \hline
        \textbf{Network Training Parameter} & \textbf{Value} \\ \hline
        Batch Size & 1000 \\ 
        Epochs & 100\\ 
        Metrics & Accuracy \\ 
    \end{tabular}
    \vspace{6pt}
    \caption{Experimental parameters.}
    \label{tab:alr_exp_parameters}
\end{table}

Two benchmarks were designed for AutoLR. Benchmark I compares the evolved optimizers with several standard adaptive optimizers; this benchmark will show us how the evolved solutions compare to the man made approaches that were developed over the years. Each of the standard optimizers was chosen to represent an advancement in the field:

\begin{itemize}
    \item Nesterov Momentum - Momentum-based optimizers.
    \item RMSprop - Discounted moving averages as a scaling factor.
    \item Adam - Bias correction.
\end{itemize}

Benchmark II uses the evolved optimizers in a different network and dataset to test their quality as out of the box optimizers. In this scenario the optimizers are compared with Adam which is informally regarded as the best optimizer prior to hyperparameter optimization. The dataset chosen for this scenario was CIFAR-10 as it is a common problem used to evaluate machine learning approaches. This dataset was not used during evolution due to its large size but since the benchmark phase requires few evaluations this is not a problem. The architecture used was the Keras CIFAR-10 architecture found in \cite{keras_cifar10}.

These benchmarks are summarized in Table~\ref{alr_setup}.

\begin{table}[ht]
\centering
\begin{tabular}{|c|c|c|c|}
\hline
\rowcolor[HTML]{EFEFEF} 
\textbf{Scenario}                                    & \textbf{Network Architecture}   & \textbf{Dataset}                & \textbf{Optimizers} \\ \hline
\cellcolor[HTML]{EFEFEF}                             &                                 &                                 & Evolved             \\ \cline{4-4} 
\cellcolor[HTML]{EFEFEF}                             &                                 &                                 & Adam                \\ \cline{4-4} 
\cellcolor[HTML]{EFEFEF}                             &                                 &                                 & RMSprop             \\ \cline{4-4} 
\multirow{-4}{*}{\cellcolor[HTML]{EFEFEF}\textbf{I}} & \multirow{-4}{*}{Keras-MNIST}   & \multirow{-4}{*}{Fashion-MNIST} & Nesterov            \\ \hline
\cellcolor[HTML]{EFEFEF}                             &                                 &                                 & Evolved             \\ \cline{4-4} 
\multirow{-2}{*}{\cellcolor[HTML]{EFEFEF}\textbf{II}} & \multirow{-2}{*}{Keras-CIFAR10} & \multirow{-2}{*}{CIFAR10}       & Adam                \\ \hline
\end{tabular}
\caption{Benchmark scenarios for Adaptive AutoLR.}
\label{alr_setup}
\end{table}


\subsubsection{Results}
Two optimizers were selected from the evolutionary runs. The first one selected was the best performing optimizer across all runs, a simplified version of this optimizer is shown in Equation~\ref{eq:sign_optimizer_simp}.

\begin{equation}
\begin{aligned}
w_{t} &= w_{t - 1} - 0,0009 * sign(\nabla l(w_{t - 1}))
\end{aligned}
\label{eq:sign_optimizer_simp}
\end{equation}

This optimizer is unusual in a few ways. The gradient is never used; only its sign. As a result this optimizer always changes the weights by a fixed amount in the direction of the gradient. We named this optimizer the \textit{Sign Optimizer}. The Sign optimizer does not exhibit any adaptive behavior; since evolving adaptive optimizers was the main object of this system we decided to also select the best adaptive optimizer for benchmark.

The best adaptive optimizer is simplified in Equation~\ref{eq:poly_optimizer_simp}. This optimizer is, as far as we know, a novel approach to adaptive optimizers. The unique aspect of this solution is the presence of a squared auxiliary variable that was not found in standard approaches. This optimizer is named Adaptive Evolutionary Squared (ADES) after its defining characteristic.

\begin{equation}
\begin{aligned}
y_{t} &= (1 - 0,08922) y_{t - 1} \\
&- (0,08922 * y_{t-1}^2 + 0,0891 y_{t - 1} \nabla l(w_{t - 1}) \\
&+ 0,0891 \nabla l(w_{t - 1}))\\
w_{t} &= w_{t - 1} + y_{t}
\end{aligned}
\label{eq:poly_optimizer_simp}
\end{equation}

In Benchmark I the evolved optimizers (Sign and ADES) are compared with standard approaches in their native machine learning task. The standard optimizers were all tested using their default parameters as found in the Keras library. The results are shown in Table~\ref{alr_results_1}. In this scenario ADES' test accuracy is close to RMSprop and Adam, surpassing Nesterov's momentum. The Sign optimizer did not perform as well. The minuscule increments performed by this optimizer do not allow the network to consistently perform as well as the others. This is evidenced by the test accuracy achieved with this optimizer as well as the comparatively large standard deviation. In sum, the results obtained with ADES in this benchmark are promising. Although the evolved optimizer did not obtain the best results it was able to remain competitive with methods that have been researched for years.

In Benchmark II the evolved optimizers are compared with Adam on a different machine learning task. The purpose of this experiment is to analyze how useful the evolved optimizers are as standalone tools removed from the framework. The results are shown in Table~\ref{alr_results_2}. The most notable result is that ADES is the best performing optimizer in this benchmark. While the difference in accuracy is not massive it is important to acknowledge that an evolved optimized is able to out perform a state of the art solution outside of its native task. This is particularly interesting when we consider that the supposed advantage of an evolved optimizer is that it is fine tuned for the task it is evolved in. The fact that ADES remains competitive when moved to other problems props up AutoLR as a tool to create general optimizers.

\begin{table}[ht]
\centering
\resizebox{\columnwidth}{!}{\begin{tabular}{|
>{\columncolor[HTML]{EFEFEF}}c |c|c|c|}
\hline
\textbf{Optimizers} & \cellcolor[HTML]{EFEFEF}\textbf{Validation Accuracy} & \cellcolor[HTML]{EFEFEF}\textbf{Test Accuracy} & \cellcolor[HTML]{EFEFEF}\textbf{Generalization Rate} \\ \hline
ADES                & 93.05 $\pm$ 0.49\%                                     & 92.45 $\pm$ 0.20\%                               & \textbf{99.36\%}                                    \\ \hline
Sign                & 91.88 $\pm$ 0.87\%                                     & 89.80 $\pm$ 0.59\%                               & 97.75\%                                             \\ \hline
Adam                & \textbf{93.40 $\pm$ 0.36\%}                            & 92.67 $\pm$ 0.12\%                      & 99.21\%                                             \\ \hline
RMSprop             & 93.34 $\pm$ 0.37\%                                     & \textbf{92.71 $\pm$ 0.19}\%                               & 99.32\%                                             \\ \hline
Nesterov            & 91.97 $\pm$ 0.36\%                                     & 90.62 $\pm$ 0.32\%                               & 99.21\%                                             \\ \hline
\end{tabular}
}
\caption{Benchmark results of evolved (ADES, Sign) and standard (Adam, RMSprop, Nesterov) optimizers in scenario I (Fashion-MNIST).}
\label{alr_results_1}
\end{table}

\begin{table}[ht]
\centering
\resizebox{\columnwidth}{!}{%
\begin{tabular}{|
>{\columncolor[HTML]{EFEFEF}}c |c|c|c|}
\hline
\textbf{Optimizers} & \cellcolor[HTML]{EFEFEF}\textbf{Validation Accuracy} & \cellcolor[HTML]{EFEFEF}\textbf{Test Accuracy} & \cellcolor[HTML]{EFEFEF}\textbf{Generalization Rate} \\ \hline
ADES                & \textbf{80.53} $\pm$ 0.81\%                                     & \textbf{79.72} $\pm$ 0.42\%                               & \textbf{99.00\%}                                    \\ \hline
Sign                & 62.89 $\pm$ 1.42\%                                     & 62.11 $\pm$ 2.18\%                               & 98.75\%                                             \\ \hline
Adam                & 79.19 $\pm$ 0.71\%                            & 78.76 $\pm$ 0.31\%                      & 99.46\%                                             \\ \hline
\end{tabular}
}
\caption{Benchmark results of evolved (ADES, Sign) and standard (Adam, RMSprop, Nesterov) optimizers in scenario II (CIFAR10).}
\label{alr_results_2}
\end{table}
\subsection{Bayesian Optimization}
In light of the results suggesting that ADES could be utilized as a general LR optimizer a final experiment was conducted. To better understand the potential of ADES we performed Bayesian optimization on the hyperparameters of all adaptive optimizers (including ADES) in all problems tackled in this work. In this experiment we are looking to understand how applicable ADES is compared to standard approaches.

Tables~\ref{task1},~\ref{task2} and \ref{task3} show the results of all four optimizers across the three tasks. For tasks I and II the ranking is the same: Nesterov's momentum performs the best, followed by ADES, Adam and RMSprop in that order. The margin are small but once again ADES is shown to be able to compete with the standard approaches. In the third task ADES showed some bizarre behavior. While this is not reflected in the final results, observed during the optimization process that certain hyperparameter values were not able to train the network in any capacity. In other words, when unoptimized ADES could produce accuracy as low as 10\% in task III. In the final results ADES was not able to perform as well as the other optimizers, perhaps because the hyperparameters were not as easily tuned.

\begin{table}[h]
\resizebox{\columnwidth}{!}{%
\begin{tabular}{|c|c|c|c|c|c|}
\hline
\rowcolor[HTML]{EFEFEF} 
Task                                        & Optimizer                  & \multicolumn{3}{c|}{\cellcolor[HTML]{EFEFEF}Parameters}                                           & Test Accuracy             \\ \hline
\cellcolor[HTML]{EFEFEF}                    &                            & \cellcolor[HTML]{EFEFEF}lr        & \cellcolor[HTML]{EFEFEF}beta1 & \cellcolor[HTML]{EFEFEF}beta2 &                           \\ \cline{3-5}
\cellcolor[HTML]{EFEFEF}                    & \multirow{-2}{*}{Adam}     & 0.004786418106                    & 0.9013530508                  & 0.9440879038                  & \multirow{-2}{*}{92.57\%} \\ \cline{2-6} 
\cellcolor[HTML]{EFEFEF}                    &                            & \cellcolor[HTML]{EFEFEF}lr        & \multicolumn{2}{c|}{\cellcolor[HTML]{EFEFEF}momentum}         &                           \\ \cline{3-5}
\cellcolor[HTML]{EFEFEF}                    & \multirow{-2}{*}{Nesterov} & 0.01241733151                     & \multicolumn{2}{c|}{0.9279183679}                             & \multirow{-2}{*}{92.84\%} \\ \cline{2-6} 
\cellcolor[HTML]{EFEFEF}                    &                            & \cellcolor[HTML]{EFEFEF}lr        & \multicolumn{2}{c|}{\cellcolor[HTML]{EFEFEF}rho}              &                           \\ \cline{3-5}
\cellcolor[HTML]{EFEFEF}                    & \multirow{-2}{*}{RMS}      & 0.001828827734                    & \multicolumn{2}{c|}{0.9750144315}                             & \multirow{-2}{*}{92.50\%} \\ \cline{2-6} 
\cellcolor[HTML]{EFEFEF}                    &                            & \cellcolor[HTML]{EFEFEF}beta1     & \multicolumn{2}{c|}{\cellcolor[HTML]{EFEFEF}beta2}            &                           \\ \cline{3-5}
\multirow{-8}{*}{\cellcolor[HTML]{EFEFEF}I} & \multirow{-2}{*}{ADES}     & \multicolumn{1}{l|}{0.9800744569} & \multicolumn{2}{l|}{0.9968261576}                             & \multirow{-2}{*}{92.69\%} \\ \hline
\end{tabular}}
\caption{Best results obtained by the four optimizers during bayesian optimization in Task I}
\label{task1}
\end{table}

\begin{table}[]
\resizebox{\columnwidth}{!}{%
\begin{tabular}{|c|c|c|c|c|c|}
\hline
\rowcolor[HTML]{EFEFEF} 
Task                                         & Optimizer                  & \multicolumn{3}{c|}{\cellcolor[HTML]{EFEFEF}Parameters}                                           & Test Accuracy             \\ \hline
\cellcolor[HTML]{EFEFEF}                     &                            & \cellcolor[HTML]{EFEFEF}lr        & \cellcolor[HTML]{EFEFEF}beta1 & \cellcolor[HTML]{EFEFEF}beta2 &                           \\ \cline{3-5}
\cellcolor[HTML]{EFEFEF}                     & \multirow{-2}{*}{Adam}     & 1.60E-03                          & 0.935380748                   & 0.9405771305                  & \multirow{-2}{*}{80.01\%} \\ \cline{2-6} 
\cellcolor[HTML]{EFEFEF}                     &                            & \cellcolor[HTML]{EFEFEF}lr        & \multicolumn{2}{c|}{\cellcolor[HTML]{EFEFEF}momentum}         &                           \\ \cline{3-5}
\cellcolor[HTML]{EFEFEF}                     & \multirow{-2}{*}{Nesterov} & 0.01392763473                     & \multicolumn{2}{c|}{0.9807391289}                             & \multirow{-2}{*}{80.84\%} \\ \cline{2-6} 
\cellcolor[HTML]{EFEFEF}                     &                            & \cellcolor[HTML]{EFEFEF}lr        & \multicolumn{2}{c|}{\cellcolor[HTML]{EFEFEF}rho}              &                           \\ \cline{3-5}
\cellcolor[HTML]{EFEFEF}                     & \multirow{-2}{*}{RMS}      & 1.83E-03                          & \multicolumn{2}{c|}{0.9750144315}                             & \multirow{-2}{*}{78.76\%} \\ \cline{2-6} 
\cellcolor[HTML]{EFEFEF}                     &                            & \cellcolor[HTML]{EFEFEF}beta1     & \multicolumn{2}{c|}{\cellcolor[HTML]{EFEFEF}beta2}            &                           \\ \cline{3-5}
\multirow{-8}{*}{\cellcolor[HTML]{EFEFEF}II} & \multirow{-2}{*}{ADES}     & \multicolumn{1}{l|}{0.9417022005} & \multicolumn{2}{c|}{0.9720324493}                             & \multirow{-2}{*}{80.61\%} \\ \hline
\end{tabular}
}
\caption{Best results obtained by the four optimizers during bayesian optimization in Task II}
\label{task2}

\end{table}

\begin{table}[]
\resizebox{\columnwidth}{!}{%
\begin{tabular}{|c|c|c|c|c|c|}
\hline
\rowcolor[HTML]{EFEFEF} 
Task                                          & Optimizer                  & \multicolumn{3}{c|}{\cellcolor[HTML]{EFEFEF}Parameters}                                           & Test Accuracy             \\ \hline
\cellcolor[HTML]{EFEFEF}                      &                            & \cellcolor[HTML]{EFEFEF}lr        & \cellcolor[HTML]{EFEFEF}beta1 & \cellcolor[HTML]{EFEFEF}beta2 &                           \\ \cline{3-5}
\cellcolor[HTML]{EFEFEF}                      & \multirow{-2}{*}{Adam}     & 1.14E-04                          & 0.9417022005                  & 0.9720324493                  & \multirow{-2}{*}{92.83\%} \\ \cline{2-6} 
\cellcolor[HTML]{EFEFEF}                      &                            & \cellcolor[HTML]{EFEFEF}lr        & \multicolumn{2}{c|}{\cellcolor[HTML]{EFEFEF}momentum}         &                           \\ \cline{3-5}
\cellcolor[HTML]{EFEFEF}                      & \multirow{-2}{*}{Nesterov} & 0.004455187854                    & \multicolumn{2}{c|}{0.9107494129}                             & \multirow{-2}{*}{92.90\%} \\ \cline{2-6} 
\cellcolor[HTML]{EFEFEF}                      &                            & \cellcolor[HTML]{EFEFEF}lr        & \multicolumn{2}{c|}{\cellcolor[HTML]{EFEFEF}rho}              &                           \\ \cline{3-5}
\cellcolor[HTML]{EFEFEF}                      & \multirow{-2}{*}{RMS}      & 1.14E-04                          & \multicolumn{2}{c|}{0.9302332573}                             & \multirow{-2}{*}{92.71\%} \\ \cline{2-6} 
\cellcolor[HTML]{EFEFEF}                      &                            & \cellcolor[HTML]{EFEFEF}beta1     & \multicolumn{2}{c|}{\cellcolor[HTML]{EFEFEF}beta2}            &                           \\ \cline{3-5}
\multirow{-8}{*}{\cellcolor[HTML]{EFEFEF}III} & \multirow{-2}{*}{ADES}     & \multicolumn{1}{l|}{0.9876389152} & \multicolumn{2}{c|}{0.9894606664}                             & \multirow{-2}{*}{91.76\%} \\ \hline
\end{tabular}
}
\caption{Best results obtained by the four optimizers during bayesian optimization in Task III}
\label{task3}
\end{table}

%% file: IEEEtran/5-Conclusion.tex
This work documents the implementation and validation of the AutoLR framework. 
This framework is capable of producing novel LR optimizers through an evolutionary algorithm. 
Two sets of experiments were performed in order to assess the viability of this approach. In these experiments the framework was given limited resources to evolve optimizers. 
During the evolutionary process there was no incentive for the solutions to imitate traditional optimizers as they were rated solely based on their performance.

The first round of experimentation focused on the development of simple dynamic LR optimizer approximations. The best optimizer evolved in these experiments showed consistently better results than the established baseline.

Since dynamic LR optimizers are only a small subset of modern learning rate optimization, we developed another, more comprehensive experimental setup. In the second set of experiments the system was able to evolve all types of LR optimizers, including the more sophisticated adaptive optimizers. One of the optimizers evolved under these circumstances was able to perform on par with the established baselines.

This optimizer, called ADES, showed other interesting properties. Despite being evolved in a specific environment this optimizer was able to out perform Adam (a standard optimizer known for its adaptability) when moved to a different task. This is notable since an evolved optimizer's supposed advantage is the opportunity to specify for the task it is evolved in.
This result prompted another test where bayesian optimization was performed on several LR optimizers (including ADES) across multiple tasks. In this test ADES was still able to perform on par with standard solution suggesting it may be used as a standalone tool. Furthermore, the results obtained with ADES indicate that the AutoLR framework can be used to create new general LR optimizers that can be used on a breadth of problems.
To summarize, the contributions of this paper are as follows:
\begin{itemize}
    \item The proposal of AutoLR, an evolutionary framework capable of producing LR optimizers.
    \item Evolution, benchmark and analysis of two types evolved optimizers: dynamic and adaptive.
    \item The discovery of ADES, the first automatically generated LR optimizer capable of competing with state of the art learning rate optimization approaches.
\end{itemize}